\documentclass[10pt,twocolumn,letterpaper]{article}

\usepackage{cvpr}              % To produce the CAMERA-READY version
\usepackage{graphicx}
\usepackage{amsmath}
\usepackage{amssymb}
\usepackage{booktabs}
\usepackage{booktabs, bbding, multirow,bbm}
\usepackage{amsmath,amssymb}
\usepackage{makecell}
\usepackage{algorithm}
\usepackage{algorithmic}

\usepackage[dvipsnames]{xcolor}

\usepackage[pagebackref,breaklinks,colorlinks]{hyperref}

\usepackage[capitalize]{cleveref}
\crefname{section}{Sec.}{Secs.}
\Crefname{section}{Section}{Sections}
\Crefname{table}{Table}{Tables}
\crefname{table}{Tab.}{Tabs.}

\def\cvprPaperID{5332} % *** Enter the CVPR Paper ID here
\def\confName{CVPR}
\def\confYear{2023}

\begin{document}
% \DeclareSubrefFormat{parens}{#1(#2)}
\title{Boosting Semi-Supervised Learning by Exploiting All Unlabeled Data}

\author{
    Yuhao Chen$^1$ \ 
    Xin Tan$^2$ \ \  
    Borui Zhao$^1$ \ 
    Zhaowei Chen$^1$ \ 
    Renjie Song$^1$\ 
    Jiajun Liang$^1$\ \ 
    Xuequan Lu$^3$\\
    $^1$MEGVII Technology \qquad
    $^2$East China Normal University \qquad 
    $^3$Deakin University\\
    {\tt\small \{yhao.chen0617, zhaoborui.gm, chaowechan\}@gmail.com, xtan@cs.ecnu.edu.cn}\\
    {\tt\small \{songrenjie, liangjiajun\}@megvii.com,  xuequan.lu@deakin.edu.au}
}

% \and
\maketitle

%%%%%%%%% ABSTRACT
\begin{abstract}
Semi-supervised learning (SSL) has attracted enormous attention due to its vast potential of mitigating the dependence on large labeled datasets. The latest methods (e.g., FixMatch) use a combination of consistency regularization and pseudo-labeling to achieve remarkable successes. However, these methods all suffer from the waste of complicated examples since all pseudo-labels have to be selected by a high threshold to filter out noisy ones. Hence, the examples with ambiguous predictions will not contribute to the training phase. For better leveraging all unlabeled examples, we propose two novel techniques: Entropy Meaning Loss (EML) and Adaptive Negative Learning (ANL). EML incorporates the prediction distribution of non-target classes into the optimization objective to avoid competition with target class, and thus generating more high-confidence predictions for selecting pseudo-label. ANL introduces the additional negative pseudo-label for all unlabeled data to leverage low-confidence examples. It adaptively allocates this label by dynamically evaluating the top-$k$ performance of the model. EML and ANL do not introduce any additional parameter and hyperparameter. We integrate these techniques with FixMatch, and develop a simple yet powerful framework called FullMatch. Extensive experiments on several common SSL benchmarks (CIFAR-10/100, SVHN, STL-10 and ImageNet) demonstrate that FullMatch exceeds FixMatch by a large margin. Integrated with FlexMatch (an advanced FixMatch-based framework), we achieve state-of-the-art performance. \textit{Source code is at https://github.com/megvii-research/FullMatch.} 
\end{abstract}

%%%%%%%%% BODY TEXT
\section{Introduction}
\label{sec:intro}

Semi-supervised learning (SSL) is proposed to leverage an abundance of unlabeled data to enhance the model's performance when labeled data is limited~\cite{ref_intro_ssl}. Consistency regularization~\cite{ref_mixmatch,ref_fake4,ref_2model} and pseudo labeling~\cite{ref_dmt, ref_entropy_min,ref_entropy_min_2} have shown significant ability for leveraging unlabeled data and thus they are widely used in SSL frameworks. Recently, FixMatch-based methods~\cite{ref_freematch, ref_fixmatch, ref_semco, ref_flexmatch} that combine the two technologies in a unified framework have achieved noticeable successes. Specifically, they apply weak-augmentation (e.g., only random flip and shift) to unlabeled data and obtain their predictions, and then corresponding one-hot pseudo-label is generated if the largest prediction confidence is beyond the predefined threshold (e.g., 0.95), finally it is used as a training target when inputting the strongly-augmented examples (e.g., RandAugment~\cite{ref_randaug}, Cutout~\cite{ref_cutout}).

\begin{figure}[t]
  \centering
  \hspace{-30pt}
  \begin{subfigure}{0.46\linewidth}
    \includegraphics[width=1.16\linewidth]{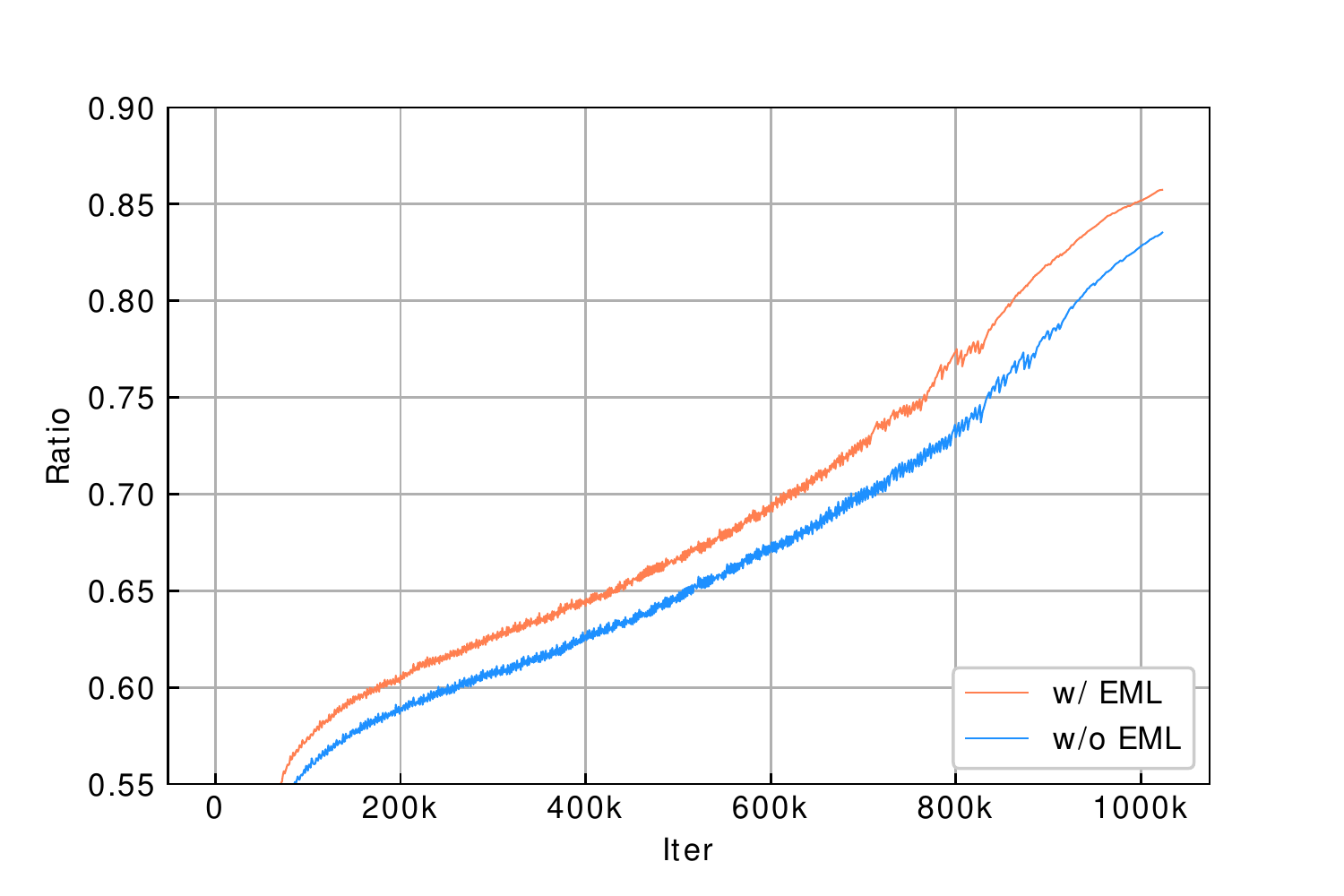}
    \caption{}
    \label{visualize of eml}
  \end{subfigure}
  \hspace{-5pt}
  \begin{subfigure}{0.46\linewidth}
    \includegraphics[width=1.16\linewidth]{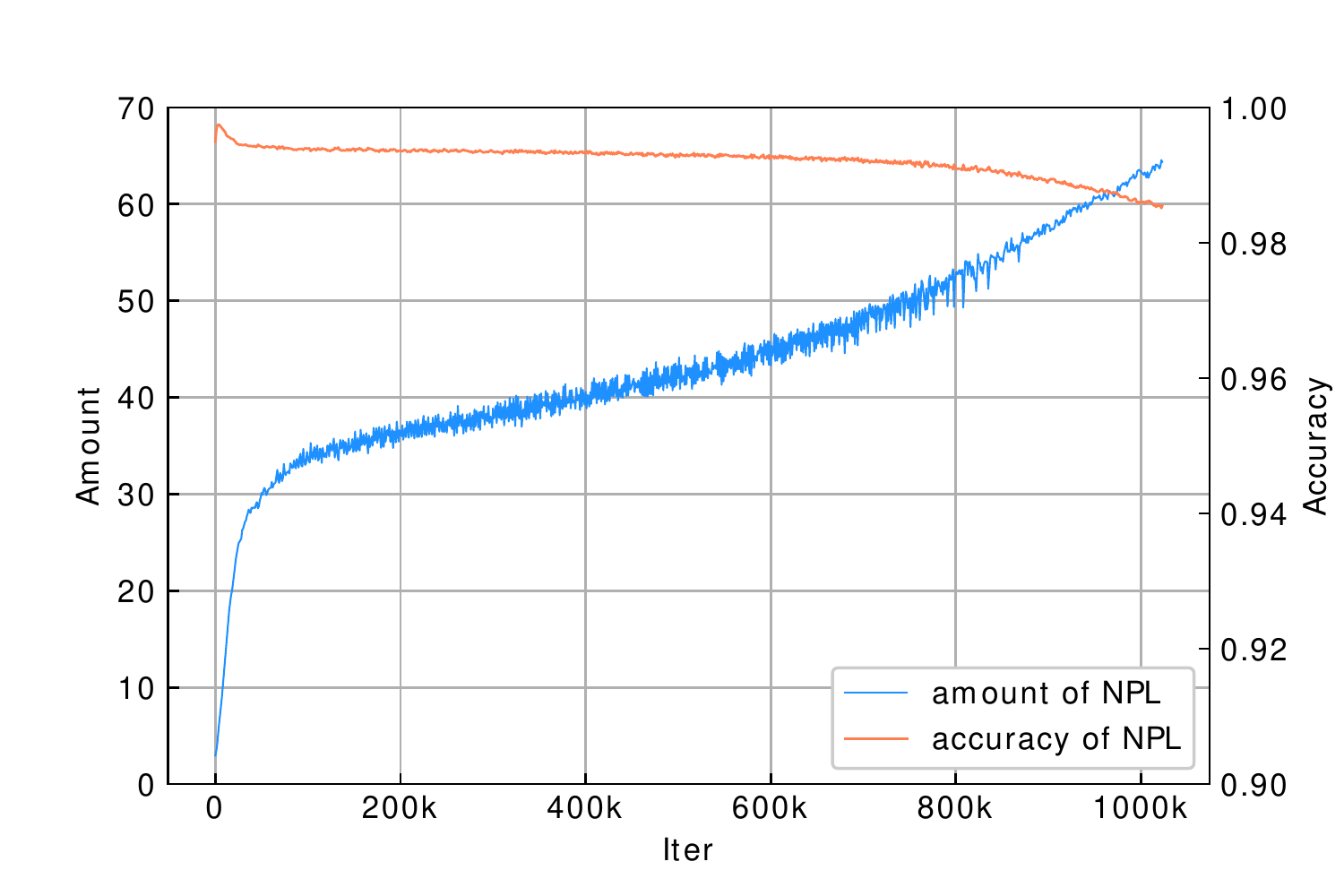}
    \caption{}
    \label{visualize of anl}
  \end{subfigure}
   \caption{Visualization of the experimental results on CIFAR-100 with 10000 labeled images. Evaluations are done every 1K iterations. (a) The increasing proportion of examples with pseudo-label when applying EML to FixMatch. (b) The number of negative pseudo-labels per sample and accuracy during the whole training process.  ``NPL" denotes negative pseudo-labels.}
   \label{visualize}
   \centering
   \vspace{-5pt}
\end{figure}

However, the FixMatch-based methods still have a significant drawback that they rely on an extremely high threshold to produce accurate pseudo-labels, which results in ignoring a large number of unlabeled examples with ambiguous predictions, especially on the early and middle training stages.  We can easily observe this phenomenon according to the blue curve in Fig.~\ref{visualize}\subref{visualize of eml}, which visualizes the proportion of samples with pseudo-labels at different training iterations when applying FixMatch~\cite{ref_fixmatch} to CIFAR-100~\cite{ref_cifar} with 10,000 labels. It shows that the ratio of selected examples with pseudo-label is around 58\% after 200k iterations and merely reaches ~84\% in the end. This motivates us to exploit more unlabeled data to boost the overall performance. 

% One intuitive solution is to~\textit{assign pseudo-label for potential examples} (i.e., the maximum confidence is close to the predefined threshold). Dash~\cite{ref_dash} and FlexMatch~\cite{ref_flexmatch} introduce a dynamic threshold strategy to select more examples with pseudo-label, yet suffering from the risk of introducing incorrect pseudo-labels.  Unlike them, we propose a novel scheme to assign more examples with pseudo-label, namely Entropy Meaning Loss (EML). For examples with pseudo-label, EML imposes additional supervision on non-target classes (i.e., classes which specify the absence of a specific label) to push their prediction close to a uniform distribution, thus preventing any classes competition with the target class and generating more high-confidence predictions.
% Fig.~\ref{visualize}\subref{visualize of eml} illustrates that FixMatch equipped with EML can select more examples with the pseudo-label during the whole training process. Since EML attempts to generate high confidence rather than tuning the threshold, it can be applied to any dynamic-threshold method to further increase the amount of examples with pseudo-label. 

One intuitive solution is to~\textit{assign pseudo-label for potential examples} (i.e., the maximum confidence is close to the predefined threshold). We argue the competition between partial classes leads to failure to produce high-confidence prediction, while the unsupervised loss of FixMatch (i.e, cross-entropy) only focus on the target class when training the examples with pseudo-label.
Therefore, we propose a novel scheme to enhance confidence on target class, namely Entropy Meaning Loss (EML). For examples with pseudo-label, EML imposes additional supervision on all non-target classes (i.e., classes which specify the absence of a specific label) to push their prediction close to a uniform distribution, thus preventing any class competition with the target class.
Fig.~\ref{visualize}\subref{visualize of eml} illustrates that FixMatch equipped with EML can select more examples with the pseudo-label during the whole training process. Since EML attempts to yield more low-entropy predictions to select more examples with pseudo-label rather than tuning the threshold, it can be also applied to any dynamic-threshold methods (e.g., FlexMatch~\cite{ref_flexmatch}, Dash~\cite{ref_dash}).

\begin{figure}[t]
\centering %cmp_confidence.pdf
\setlength{\abovecaptionskip}{-2pt}
\includegraphics[width=0.9\linewidth]{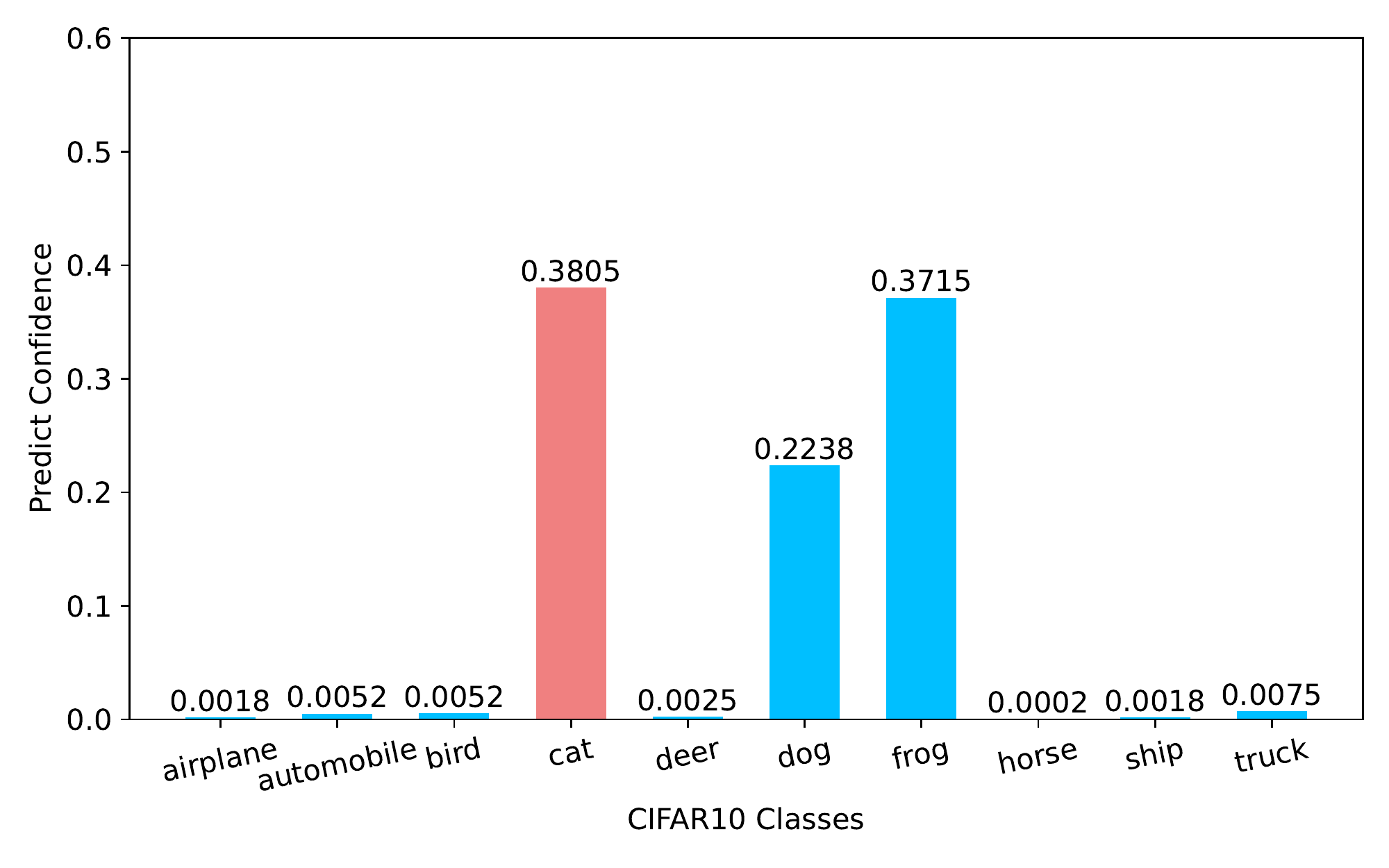}
\caption{An example of inference result of FixMatch. It can conclude that the input does not belong to these low-rank classes, such as airplanes, horse.} 
\vspace{-5pt}
\label{fixmatch_conf}
\end{figure}

Nevertheless, it is still impossible to leverage all unlabeled data by generating pseudo-labels with a threshold strategy. This motivates us to further consider~\textit{how to utilize the low-confidence unlabeled examples without pseudo-label} (i.e., the maximum confidence is far from the predefined threshold). Intuitively, the prediction may get confused among the top classes, but it will be confident that the input does not belong to the categories ranked after these classes.
Fig.~\ref{fixmatch_conf} shows an inference result of FixMatch. The ground truth is ``cat", FixMatch is confused by several top classes (e.g., ``dog'', ``frog'') and make low-confidence prediction, however it shows highly confidence that some low-rank classes (e.g.,``airplane'', ``horse'') are not ground truth class, thus we can safely assign \textit{negative pseudo-labels} to these classes.
Based on this insight, we propose a novel method named Adaptive Negative Learning (ANL). 
%ANL adaptively allocates negative pseudo-label by dynamically assessing the model's top-$k$ performance.  
Specifically, ANL first calculate a $k$ \textit{adaptively} based on the prediction consistency, so that the accuracy of top-$k$ is close to 1, and then regard the classes ranked after $k$ as negative pseudo-labels. Furthermore, if the example is selected a pseudo-label, ANL will shrink the range of non-target classes (i.e., EML only needs constrain the top-$k$ classes except target class).
Note that ANL is a threshold-independent scheme and thus can be applied on \textit{all} unlabeled data.
%regardless of whether the prediction is a low entropy. 
As shown in Fig.~\ref{visualize}\subref{visualize of anl}, our ANL's rendered negative pseudo-labels are increasing as the model is optimized while keeping high accuracy.
In summary, our method \textbf{makes full use} of the unlabeled dataset, which is hardly ever seen in modern SSL algorithms.
% Nevertheless, it is still impossible to leverage all unlabeled data by generating pseudo-labels with a threshold strategy. This motivates us to further consider~\textit{how to utilize the low-confidence unlabeled examples without pseudo-label} (i.e., the maximum confidence is far from the predefined threshold). In this paper, we propose a novel method named Adaptive Negative Learning (ANL). ANL adaptively allocates Negative Pseudo-Label (i.e., input image does not belong to this label with a high probability) for all unlabeled data by dynamically assessing the model's top-$k$ performance. As shown in Fig.~\ref{visualize}\subref{visualize of anl}, our ANL's rendered negative pseudo-labels are increasing as the model is optimized. Since ANL allows the model to learn from low-confidence examples without (positive) pseudo-label, the model can be converged with higher accuracy.  In summary, our method \textbf{makes full use} of the unlabeled dataset, which is hardly ever seen in modern SSL algorithms.

To demonstrate the effectiveness of the proposed EML and ANL, we simply integrate EML and ANL to FixMatch and exploit a new framework named FullMatch. 
% FullMatch does not introduce any additional parameters (hyperparameters or trainable parameters). 
% We conduct experiments with various label amounts on several popular SSL benchmarks, i.e., CIFAR-10/100~\cite{ref_cifar}, SVHN~\cite{ref_svhn}, STL-10~\cite{ref_stl10} and ImageNet~\cite{ref_imagenet}. 
We conduct various experiments on CIFAR-10/100~\cite{ref_cifar}, SVHN~\cite{ref_svhn}, STL-10~\cite{ref_stl10} and ImageNet~\cite{ref_imagenet}. 
The results demonstrate that FullMatch surpasses the performance of FixMatch with a large margin while the training cost remains similar to FixMatch. Moreover, our method can be easily adapted to other FixMatch-based algorithms and obtain further improvement. For example, by combining it with FlexMatch~\cite{ref_flexmatch}, we achieve state-of-the-art performance. To summarize, our key contributions include:

% 1)  We introduce a novel Entropy Meaning Loss (EML) to select more samples with pseudo-labels. EML generates more confident and robust predictions by enforcing the non-target classes to distribute uniformly by learning additional supervision. 
1) We introduce an additional supervision namely Entropy Meaning Loss (EML) when training examples with pseudo-label, which enforces a uniform distribution of non-target classes to avoid them competition with target class and thus producing more high-confidence predictions.

2) We propose the Adaptive Negative Learning (ANL), a dynamic negative pseudo-labels allocation scheme, which renders negative pseudo-labels with very limited extra computational overhead for all unlabeled data, including the low-confidence ones. 

3) We design a simple yet effective framework named FullMatch by simply integrating FixMatch with the proposed EML and ANL, which leverages \textbf{all unlabeled data} and thus achieving remarkable gains on five benchmarks. Furthermore, our method is shown to be orthogonal to other FixMatch-based frameworks. Specifically, FlexMatch with our method, achieves state-of-the-art results.

\section{Related Work}
Semi-supervised learning is one of the fundamental tasks in machine learning and computer vision. The goal of SSL is to learn from unlabeled samples with the guidance of limited labeled samples. In this section, we focus only on approaches closely relevant to our method. 

\textbf{Entropy minimization.} It has been proved effective in SSL.~\cite{ref_entropy_min} pointed out that unlabeled data should be used to facilitate a network for generating predictions far from the decision boundary. We can achieve this by promoting the network to make low entropy output distribution (i.e., high confidence prediction) on unlabeled data.~\cite{ref_entropy_min,ref_vat} added an explicit loss term to constrain the prediction entropy across $C$ classes on all unlabeled data: $-\sum_{c=1}^{C}{\mu_{c}log(\mu_{c})}$.
% and applied a small softmax temperature to have a low entropy when calculating the target distribution of unlabeled samples.  
Pseudo-labeling~\cite{ref_pseudo_label} implicitly minimized the prediction entropy by converting model predictions to one-hot label and only retaining those when the highest class prediction probability is above on predefined threshold. Nowadays, pseudo-labeling has always been used in modern SSL algorithms \cite{ref_st++, ref_densetea,ref_selfmatch, ref_u2pl} as a component of their pipeline to produce better performance. Inspired by them, our method attempts to obtain separable classification boundaries but imposes additional supervision on non-target classes. 

\textbf{Consistency regularization.} It has been extensively used in SSL~\cite{ref_boostmis,ref_vat,ref_mixmatch} to hold similar output distribution when input was  perturbed. 
% This was first proposed in~\cite{ref_cr_1}, which employs a variety of perturbations for unlabeled data such as random max pooling and dropout~\cite{ref_dropout}. 
% Mean Teacher~\cite{ref_mean_teachers} introduces a ``teacher model'' by reloading the previously learned parameters and then constrains the output distribution at the latest epoch. 
Since data augmentation~\cite{ref_mixup,ref_cutmix,ref_fmix} shows huge superiority in supervised learning, recent works have begun to give more attention to data augmentation perturbation and have achieved significant successes. For example, UDA~\cite{ref_uda}, RemixMatch~\cite{ref_remixmatch} and FixMatch~\cite{ref_fixmatch} all employed weakly-augmented strategies to produce the training target for unlabeled data and enforce prediction consistency against strongly-augmented version. The difference between FixMatch and UDA/RemixMatch is that FixMatch adopts pseudo-labeling instead of employing a ``soft'' label by sharpening the predicted distributions, which is beneficial to entropy minimization. Therefore, FixMatch obtains better performance and is a milestone algorithm in SSL. However, the predefined threshold used in FixMatch causes certain low-confidence examples to make no contributions to the model learning, and FlexMatch introduces Curriculum Pseudo Labeling to dynamically adjust the confidence threshold and remarkably boosts performance. By contrast, the proposed EML enhances the prediction confidence of the model itself by enlarging the distinction between the target and non-target classes, thus generating more low entropy predictions under the same threshold. This reveals that our method can obtain better improvement either with a fixed threshold or dynamic threshold.

\textbf{Negative learning.} It is an indirect learning strategy where the category of the inputs is not the same as the supervised learning. Compared with the ``positive label'' (i.e., the image belongs to this category), the superiority of negative labels is less-cost and more accurate. 
% Originally, the negative learning method was proposed to learn with noisy labels in~\cite{ref_NLNL}. 
% UPS~\cite{ref_in_defense} and NS$^3$L~\cite{ref_n3l} introduce negative learning into SSL by setting a predefined small threshold. Specifically, UPS pre-defines two thresholds $\tau_p$ (e.g., 0.95) and $\tau_n$ (e.g., 0.05). If a network is sufficiently confident of a class's prediction (i.e., $p^i_c \geq \tau_p$), then the positive pseudo-label is selected; if a confidence score is extremely low (i.e., $p^i_c \leq \tau_n$), the negative pseudo-label is assigned. UPS employs an extremely small threshold $\tau_n$ (e.g., 0.05) to decrease the noise of negative pseudo-label, yet leading to the fact that only 88\% samples without pseudo-labels can be assigned negative pseudo-labels. Our proposed ANL adaptively assigns negative pseudo-labels for $all$ unlabeled data while maintaining simplicity (i.e., no extra threshold hyperparameter).
UPS~\cite{ref_in_defense} and NS$^3$L~\cite{ref_n3l} select negative labels for the classes whose probability values fall below a fixed small threshold (e.g., 0.01). In other words, these methods still utilize the ``high-confidence" prediction based on the (low) probability value. Obviously, these methods can not label samples with negative labels when given ambiguous predictions (e.g., all probability values are between 0.01 and 0.95). By contrast, our proposed ANL focus on the rank of categories rather than the probability value, it adaptively assigns negative pseudo-labels for $all$ unlabeled data while maintaining simplicity (i.e., no extra threshold hyperparameter).

\section{Method}

As discussed above, we will concentrate on two questions in this section: 1) how to allocate more examples with pseudo-label; 2) how to learn knowledge from unlabeled examples with ambiguous predictions.
% how to learn knowledge from the low-confident unlabeled samples without pseudo-labels
Correspondingly, we propose two novel and efficient techniques: Entropy Meaning Loss (EML) and Adaptive Negative Learning (ANL). EML constrains the output distribution of non-target classes to obtain more separable decision boundaries, thus generating more high-confidence predictions. ANL dynamically assigns negative pseudo-label based on the model's optimization status to leverage examples with ambiguous prediction. 
By applying these two key components to  FixMatch~\cite{ref_fixmatch}, we can employ the total unlabeled dataset and bring improvements for various baselines. In this section, we first review the key components of FixMatch. Then, we explain the proposed Entropy Meaning Loss (EML) and Adaptive Negative Learning (ANL),  respectively. Finally, we introduce the FullMatch algorithm by integrating our method with FixMatch.

\subsection{Preliminaries}
Consistency regularization is proved very useful in SSL. The original consistency loss in SSL is a $L-2$ loss. 
\begin{equation}
\sum_{i=1}^{B}(\left\| p_{m}(y|\omega(\mu^{(i)})-p_{m}(y|\phi(\mu^{(i)}))  \right\|_2^2)
\end{equation}
where $p_{m}$ denotes the prediction distribution of the model. $\omega$ and $\phi$ are different perturbations imposed on the unlabeled examples $\mu^{(i)}$. 
$B$ represents the batch size of unlabeled examples. FixMatch  introduces Pseudo-Labeling techniques related to entropy minimization in the consistency regularization process. The improved consistency loss function in FixMatch can be formulated as:  
\begin{equation}
\frac{1}{B}\sum_{i=1}^{B} \mathbbm{1}(max(Q^{(i)})\geq \tau)H(\hat{Q^{(i)}}, P^{(i)}) \label{cal unsup_loss}
\end{equation} 
where $Q^{(i)}=p_{m}(y|\omega(\mu^{(i)}))$  and $P^{(i)}=p_{m}(y|\phi(\mu^{(i)}))$ represents the prediction distribution of the weakly-augmented version and strongly-augmented version, respectively. $\omega$ and $\phi$ denote weakly and strongly augmentations. $\hat{Q^{(i)}}=argmax(Q^{(i)})$ is the hard target.  $\tau$ is a confidence threshold and $H$ represents the cross-entropy loss function. FixMatch generates the pseudo-label according to the output distribution with weak-augmented inputs, and then calculates the difference from strongly-augmented inputs.

As suggested by previous research, a high confidence threshold $\tau$ generates accurate pseudo-label but filters out lots of unlabeled data with low-confidence predictions, thus causing the under-exploration of the unlabeled data (see Fig.~\ref{visualize}\subref{visualize of eml}). We will propose two simple yet effective schemes to address the dilemma below. 

\subsection{Entropy Meaning Loss}\label{sec_eml}

We propose Entropy Meaning Loss to allocate more samples with pseudo-labels. Most current works focus on dynamically adjusting the threshold (e.g., FlexMatch, Dash). Unlike them, we aim to strengthen the representation ability of the model itself to produce more predictions far from the decision boundary (i.e., high-confidence predictions), which means it is orthogonal to those dynamic thresholding works. 

We assume $Q^{(i)}=[q^{(i)}_1,...,q^{(i)}_C]$ represents the prediction vector for the weakly-augmented version of sample $i$. Let $S^{(i)}=[s^{(i)}_1,...,s^{(i)}_C] \subseteq{\{0,1\}^{C}}$ be a binary vector denoting the selected labels, where $s^{(i)}_c=1$ represents the class $c$ that is selected as a target class (e.g., pseudo-label class) and $s^{(i)}_c=0$ when this class is absence of a specific label. The vector can be computed as: 
\begin{equation}
s^{(i)}_{c} = \mathbbm{1}(q^{(i)}_{c} \geq \tau)
\end{equation}
where $\tau$ is the selection threshold. Furthermore, we can calculate the vector $U^{(i)}=[u^{(i)}_1,...,u^{(i)}_C]$, where $u^{(i)}_c=1$ 
denotes class $c$ is a non-target class and  sample $i$ is assigned 
a pseudo-label, which is formulated as:  
% $q^{(i)}_{c}$ is the confidence of the class $c$ for a sample $i$ among the weakly-augmented inputs, and 
\begin{equation}
u^{(i)}_{c} = \mathbbm{1}(max(Q^{(i)})\geq \tau) \cdot \mathbbm{1}(s_c^{(i)}=0)
\end{equation}
% where $q^{(i)}$ is the prediction distribution of the weakly-augmented sample $i$. 

\begin{figure}[t]
\centering %cmp_confidence.pdf
\includegraphics[width=0.9\linewidth]{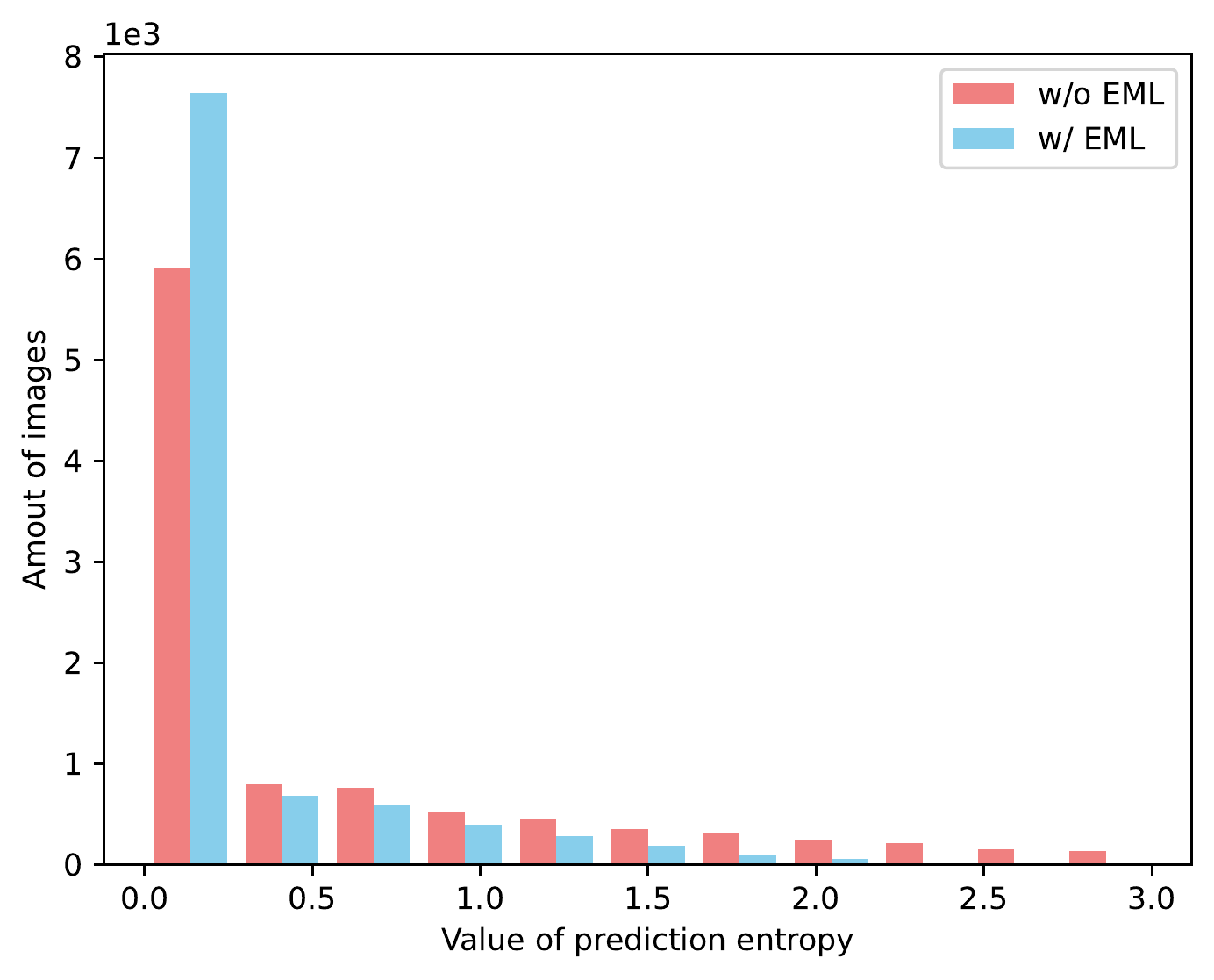}
\caption{Visualization the distribution of prediction entropy when adopting EML to FixMatch on CIFAR-100 testset . The model supervised by EML can generate more low-entropy predictions and thus select more examples with pseudo-label.} 
\vspace{-5pt}
\label{cmp_confidence}
\end{figure}

We assume $P^{(i)}=[p^{(i)}_1,...,p^{(i)}_C]$ represents the prediction confidence vector on the strongly-augmented of sample $i$, and $p^{(i)}_{tc}$ denotes the confidence of the target class (i.e., pseudo-label class). With the optimization by the unsupervised loss function (i.e, cross-entropy),
%During the optimization phase supervised by the loss function,
$p^{(i)}_{tc}$ will gradually converge to the given label (i.e., the confidence of pseudo-label class will gradually increase to 1 with model learning). But for certain challenging examples, the competition between confusion classes and target class always leads to a failure in generating high-confidence prediction. To tackle this issue, we impose an additional constraint on the rest of the categories (i.e., all non-target classes) to allow them to share the remaining confidence $1-p^{(i)}_{tc}$ equally to avoid any class competition with the target class. This can be formulated as: 
\begin{equation}
y_{c}^{(i)} = 
\frac{1-\mathbbm{1}(u^{(i)}_c=0)\cdot p^{(i)}_c}{\sum_c{\mathbbm{1}(u^{(i)}_c=1)}} \label{cal confused labels}
\end{equation}
where $y_{c}^{(i)}$ is the label of the non-target classes. It indicates that once the predicted probability of the target class is determined, other non-target classes should share the remaining confidence scores equally. 
Note that EML is just applied on the pseudo-label samples, which means $\sum_c{\mathbbm{1}(u^{(i)}_c=1)}$ is always larger than 0 due to $max(Q^{(i)})\geq \tau$.
Since $y_{c}^{(i)}$ ranges from 0 to 1, the model can be trained with a binary cross entropy (BCE) loss. Thus, our proposed Entropy Meaning Loss (EML) can be defined as:
\begin{equation}
\begin{split}
    \mathcal{L}_{eml} = -\frac{1}{BC} \sum_{i=1}^{B} \sum_{c=1}^{C} u^{(i)}_{c} \cdot [y_{c}^{(i)}log(p_c^{(i)})\\
    + (1-y_{c}^{(i)})(log(1-p_c^{(i)}))] \label{cal em}
\end{split}
\end{equation}

\begin{figure}[t]
\centering
    \leftline{
        \includegraphics[width=4.3cm,height=3.2cm]{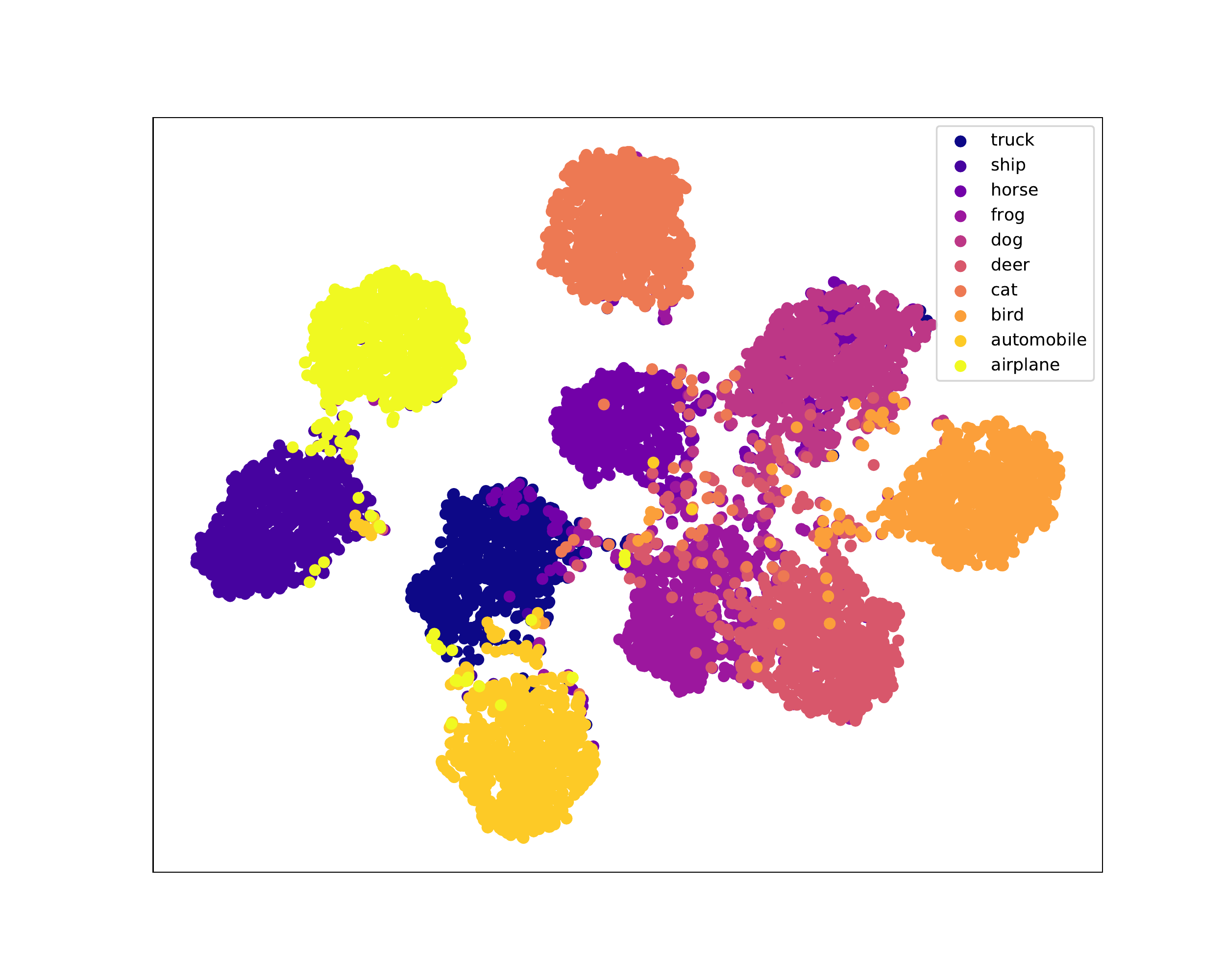}
        \hspace{-10pt}
        \includegraphics[width=4.3cm,height=3.2cm]{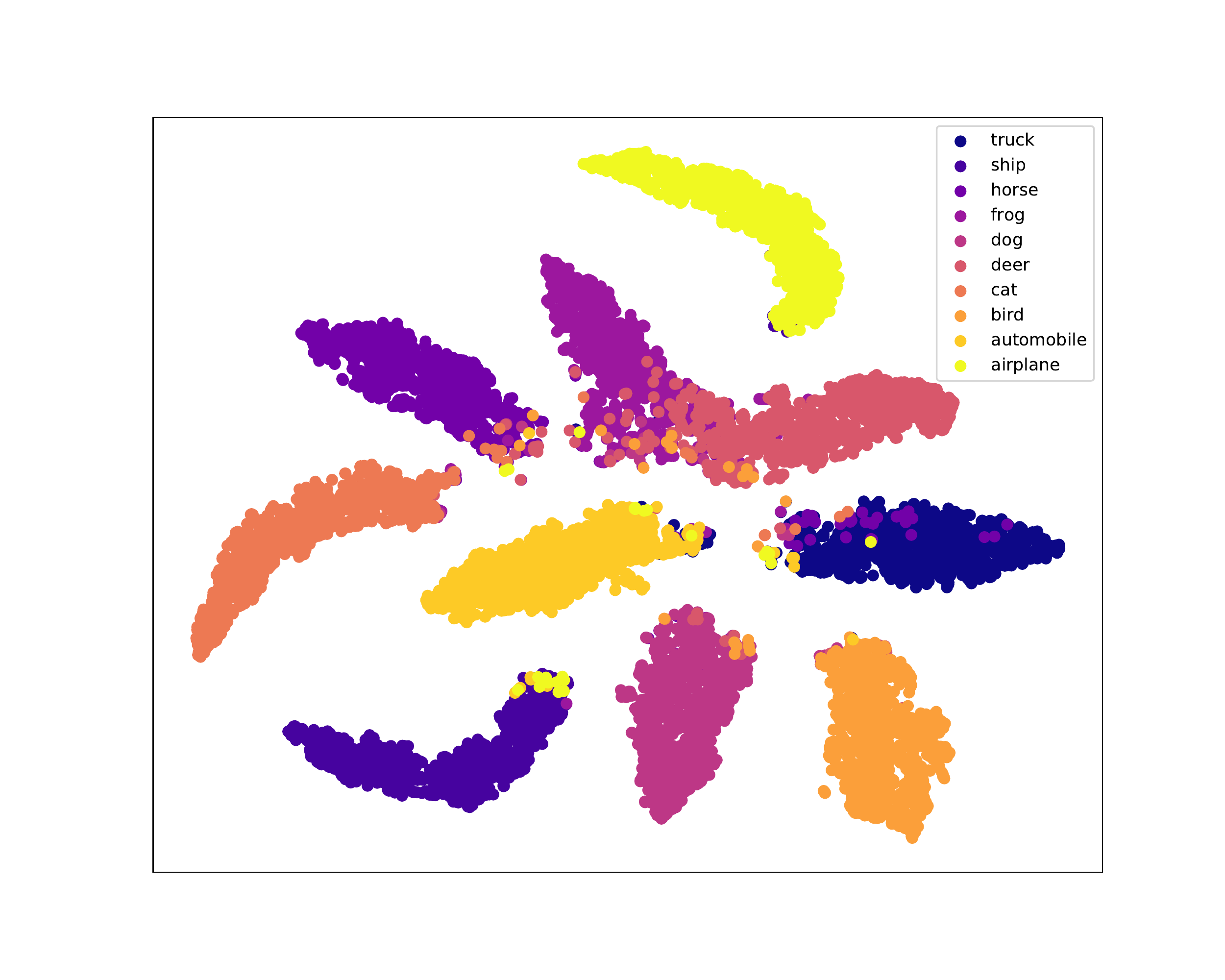}
    }
    \vspace{-6pt}
    \centerline{ {(a) CIFAR-10}}
    \vspace{-2pt}
    \leftline{
        \includegraphics[width=4.3cm,height=3.2cm]{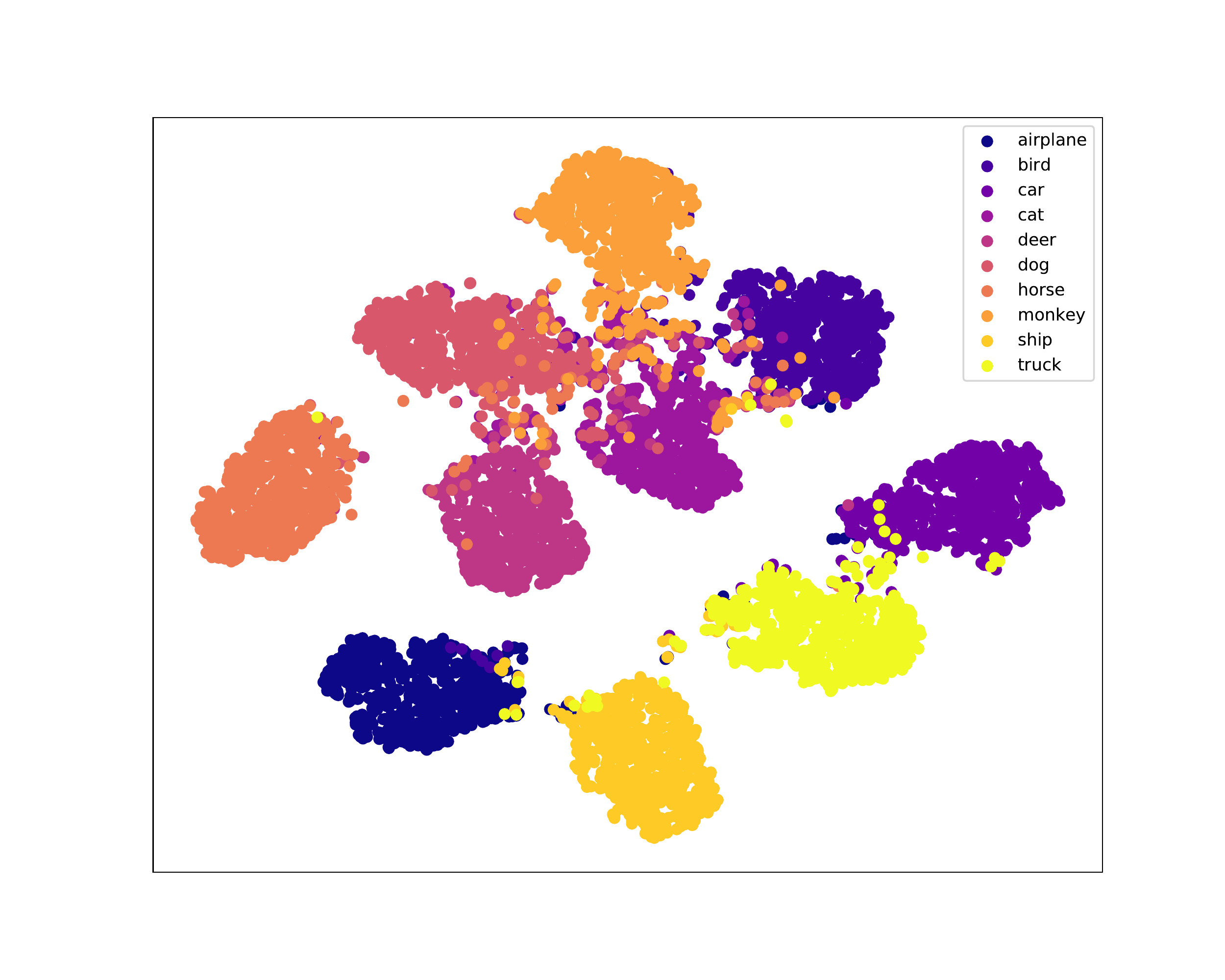}
        \hspace{-10pt}
        \includegraphics[width=4.3cm,height=3.2cm]{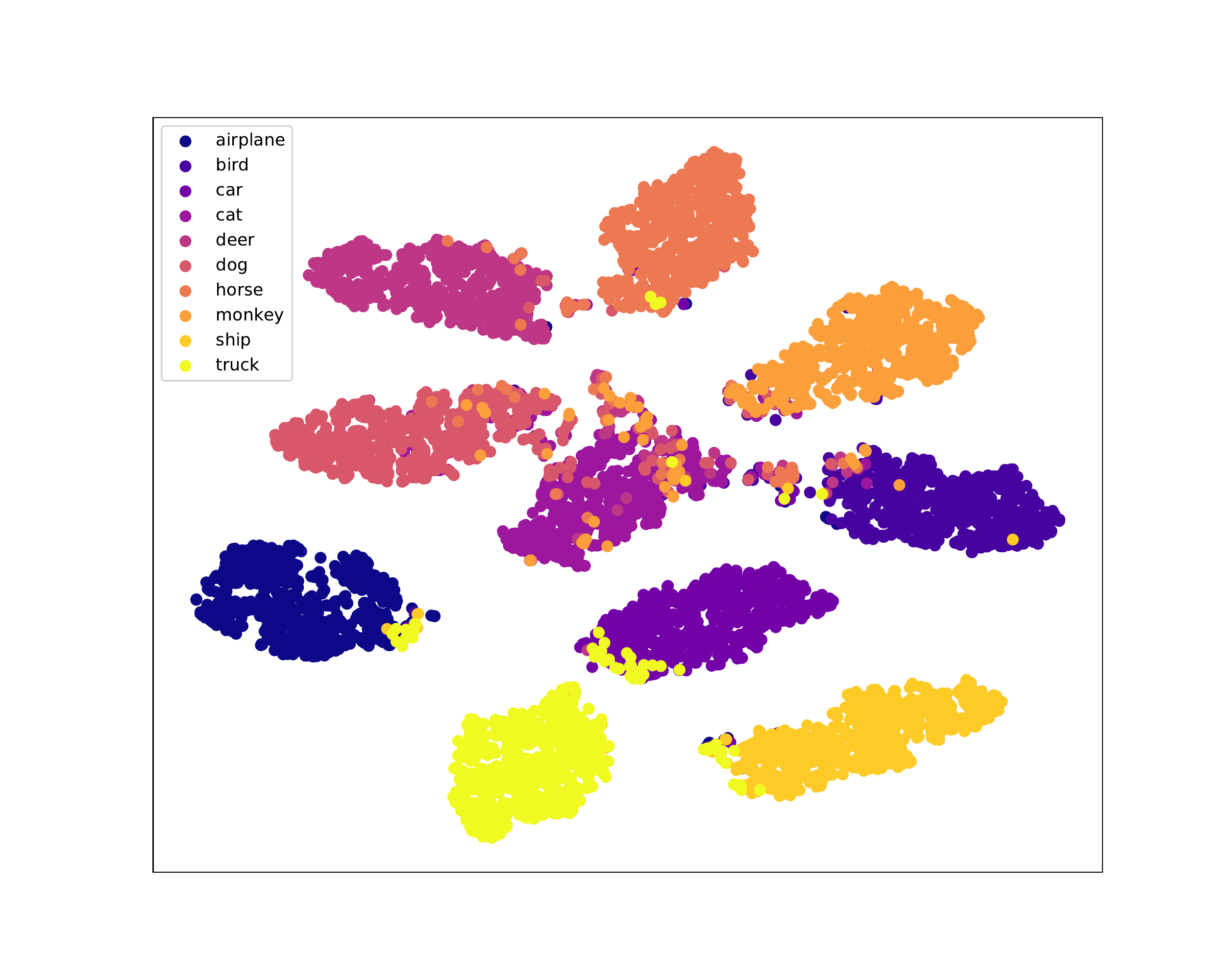}
    }
    \vspace{-5pt}
    \centerline{{(b) STL-10}}
    \caption{T-SNE visualization when adopting EML to FixMatch. The model supervised by EML (right) produces more clean and separable decision boundaries.} 
    \label{tsne}
\end{figure}

Note that $y_{c}^{(i)}$ is calculated by the score of target class, therefore EML will also produce gradient to target class, which can be computed as:
\begin{equation}
    g^{(i)}_{tc} = - \frac{1}{BC(C-1)} log(\frac{\prod \limits_{c=0,c\neq tc}^C (1-p^{(i)}_{c})}{\prod \limits_{c=0,c\neq tc}^C p^{(i)}_{c}})
\label{cal tc grad}
\end{equation}
the gradient directions of EML and unsupervised loss $\mathcal{L}_{us}$ (Eq.(~\ref{cal unsup_loss})) (i.e., cross-entropy loss) are the same, which indicates EML can cooperate with $\mathcal{L}_{us}$ to further promote the confidence of target class while constraining the distribution of the non-target classes.
% can promote the target class confidence while constraining the distribution of the non-target classes. 
For detailed proof, please refer to~\textit{Supplementary Material}, Section A.

% To intuitively illustrate the effectiveness of EML, Fig.~\ref{cmp_confidence} compares the inference results with and without introducing EML. This sample is from CIFAR-10 and the ground truth is ``cat". FixMatch only supervised by the cross-entropy loss is confused by the ``dog'' and ``frog'', resulting in missing high-confidence outputs and even making the wrong predictions. However, EML suppresses the probability of confusion categories (i.e., ``dog'' and ``frog'') to avoid the competition with   ``cat'', thus making low-entropy prediction.
To intuitively illustrate the effectiveness of EML, Fig.~\ref{cmp_confidence} compares the distributions of prediction entropy on CIFAR-100 testset with or without introducing EML. The total images is 10000. Obviously, the amount of low-entropy prediction (e.g., the value of prediction entropy is less than 0.25) will increase about 18\% (78\% vs 60\%) when introducing EML. 
We further show the effectiveness of EML by using t-SNE~\cite{ref_tsne} on CIFAR-10 and STL-10~\cite{ref_stl10}. Fig.~\ref{tsne} demonstrates that EML can produce more clean and separable boundaries, hence giving more high-confidence predictions. 

\begin{figure*}[t]
\centering
\includegraphics[width=0.9\textwidth]{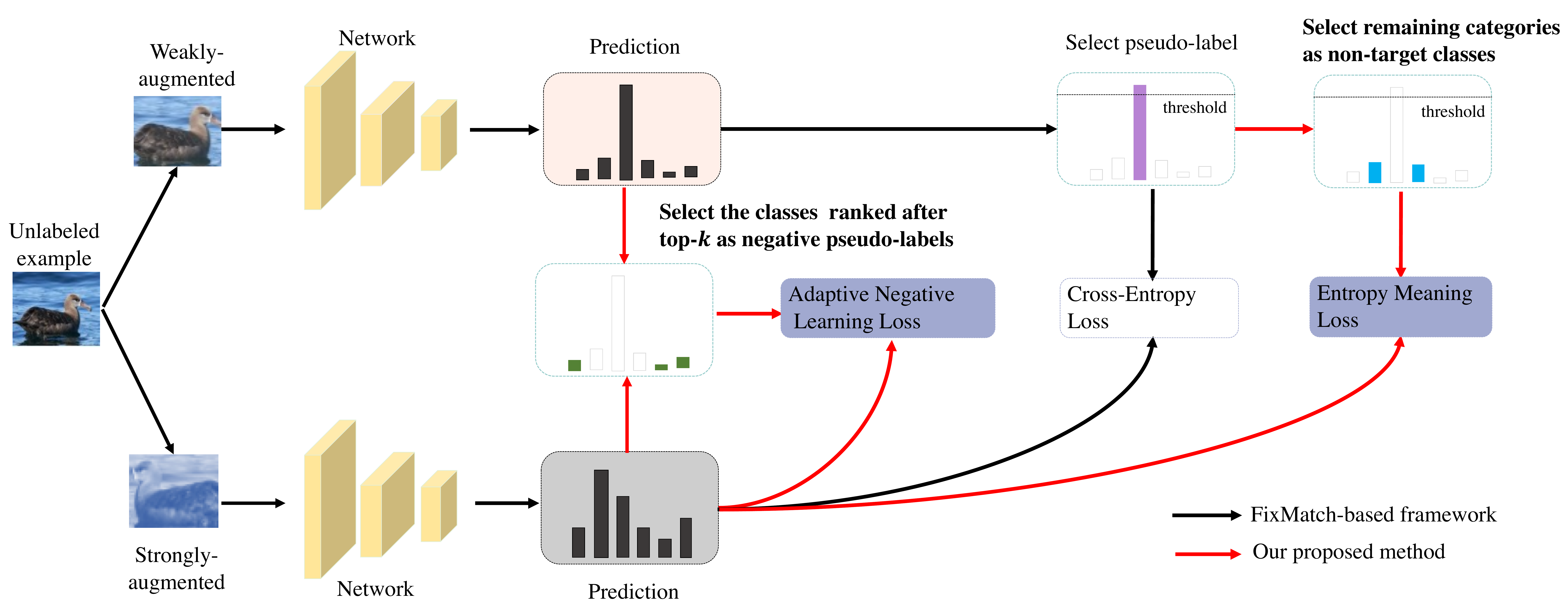}
\caption{Overview of the proposed FullMatch. First, we allocate the negative pseudo-label (green bar) for all unlabeled data with the proposed Adaptive Negative Learning. Then, if the highest probability is above the predefined threshold (dotted line), we will assign the pseudo-label (purple bar) just like FixMatch, but we optimize further remaining non-target classes (blue bar) via the proposed Entropy Meaning Loss. {The black line indicates the existing FixMatch-based methods, and the red line is our proposed method. (Best viewed in color). } } \label{FullMatch}
\vspace{-5pt}
\end{figure*}

% \begin{figure*}[t]
% \begin{minipage}[h]{0.45\linewidth}
%     \hspace{-20pt}
%     \setlength{\abovecaptionskip}{-0.03cm}
%     \includegraphics[width=8cm, height=7cm]{image/cmp_confidence.pdf}
%     \caption{\small{Visualization on CIFAR-10 inference result when adopting EML to FixMatch. The model supervised by EML generates more confident prediction distributions.} }
%     \label{cmp_confidence}
% \end{minipage}
% \hspace{4pt}
% \begin{minipage}[h]{0.45\linewidth}
%     \vspace{12pt}
%     \leftline{
%         \includegraphics[width=3.3cm,height=2.8cm]{image/fixmatch_tsne_cifar10.pdf}
%         \hspace{-10pt}
%         \includegraphics[width=3.3cm,height=2.8cm]{image/fullmatch_tsne_cifar10.pdf}
%     }
%     \vspace{-6pt}
%     \centerline{\small {(a) CIFAR-10}}
%     \vspace{-2pt}
%     \leftline{
%         \includegraphics[width=3.3cm,height=2.8cm]{image/fixmatch_tsne_stl.pdf}
%         \hspace{-10pt}
%         \includegraphics[width=3.3cm,height=2.8cm]{image/fullmatch_tsne_stl.pdf}
%     }
%     \vspace{-6pt}
%     \centerline{\small {(b) STL-10}}
%     \setlength{\abovecaptionskip}{-0.01cm}
%     \caption{\small{T-SNE visualization when adopting EML to FixMatch. The model supervised by EML (right) produces more clean and separable decision boundaries.} }
%     \label{tsne}
% \end{minipage}
% \end{figure*}

\subsection{Adaptive Negative Learning}

Since it is easy to produce ambiguous predictions on complicated scenarios (e.g., the largest confidence is only 0.3 while the threshold is 0.95), these examples are difficult to be assigned pseudo-label (filtered by the threshold or incorrect predictions), resulting in no contributions to the model optimization. To address this, we allocate an additional label with less noise to leverage these examples. 

\begin{figure}[t]
\centering
\hspace{-5pt}
\vspace{-10pt}
\begin{minipage}[h]{0.45\linewidth}
\centering
\includegraphics[width=4.3cm]{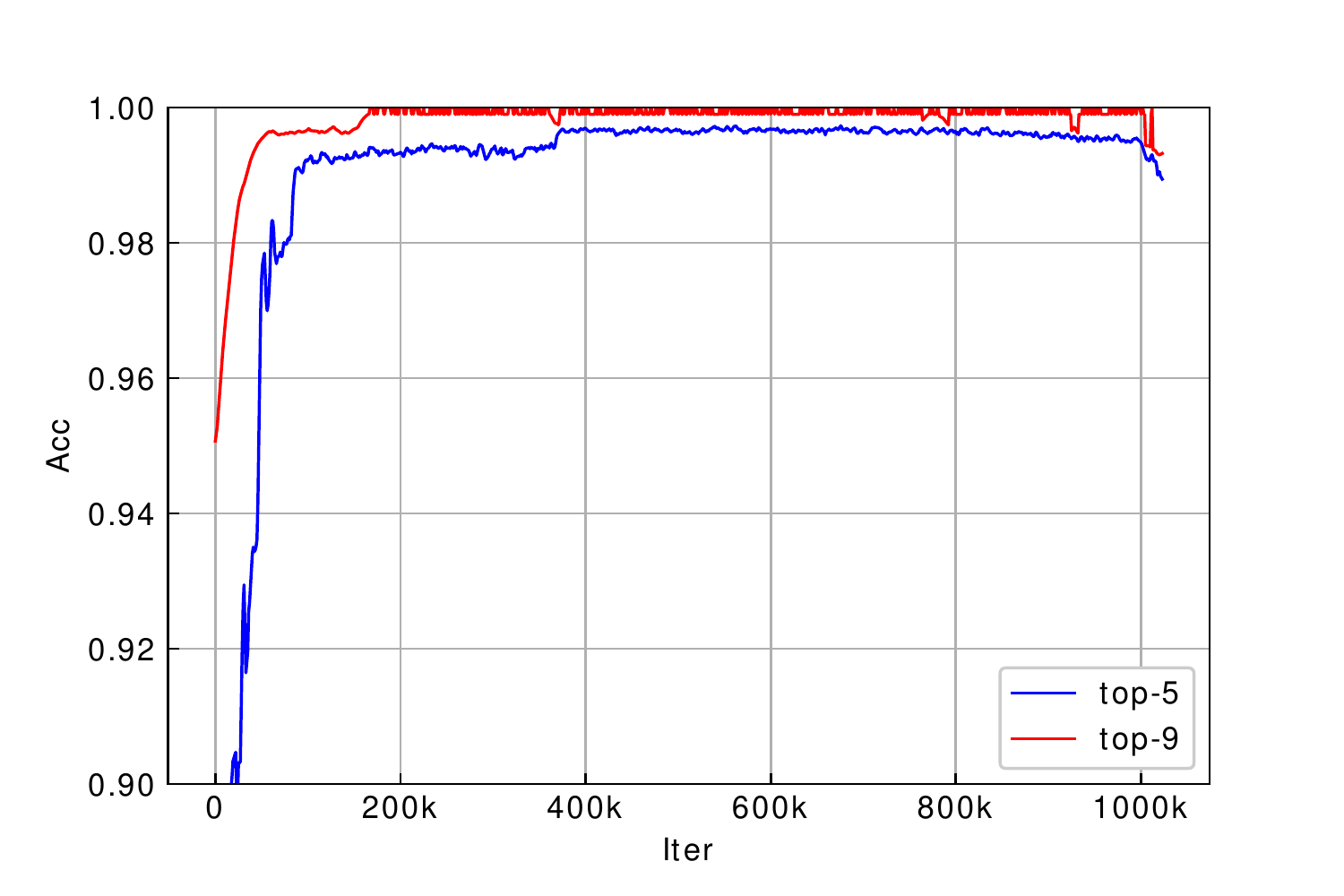}
\caption{The top-5 and top-9 accuracy curves of FixMatch during training on CIFAR-10 with 40 label samples.}
\label{topk_acc}
\end{minipage}
\hspace{8pt}
\begin{minipage}[h]{0.45\linewidth}
\centering
\includegraphics[width=4.3cm]{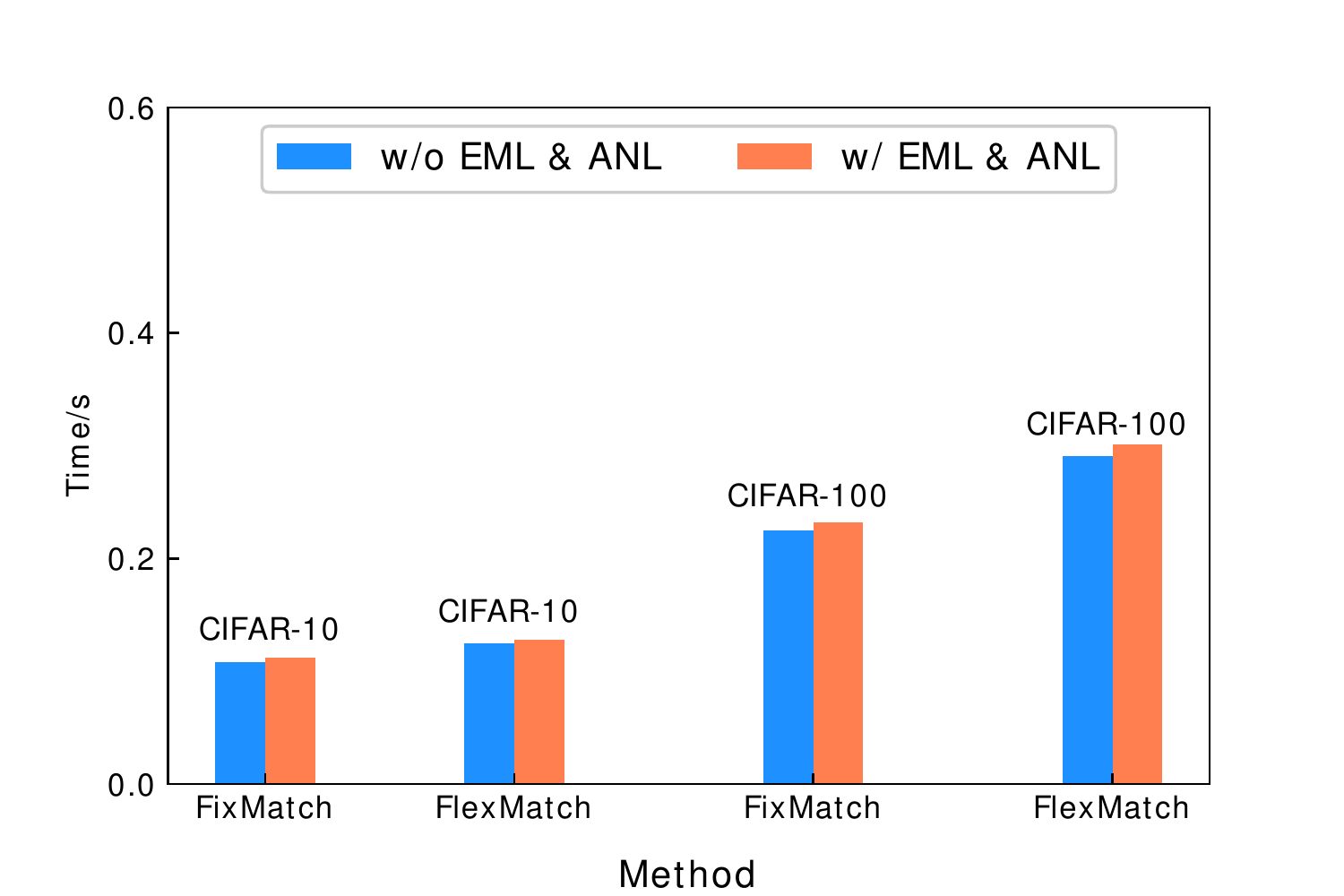}
\caption{Average training time cost of one iteration on the same Geforce RTX 2080 Ti GPU. }
\label{inference_time}
\vspace{-5pt}
\end{minipage}
\end{figure}

Fig.~\ref{topk_acc} illustrates the top-$k$ (e.g., top-5 and top-9) accuracy curves of FixMatch on CIFAR-10 when the amount of labeled data is only 40. The top-5 prediction can reach a promising accuracy after 100k iterations, which means all unlabeled data in CIFAR-10 do not belong to the $last$-5 prediction classes (i.e., categories after top-5) with a high chance when iterations are greater than 100k. This phenomenon motivates us to render negative pseudo-labels for unlabeled data. %Similarly, in the early stage of the training process, we can utilize a larger $k$ (e.g., top-9) to generate negative pseudo-labels. 

Consequently, an ideal approach is to exploit an additional dataset to evaluate the top-$k$ performance, thereby calculating a suitable $k$ value so that the top-$k$ error rate is close to zero. Since we cannot employ the test set in the model training procedure, an optional strategy is to separate an additional validation set from the labeled data. Nevertheless, this brings two severe defects: 1) separating a validation set from a labeled training set is expensive, especially when the amount of labeled data is particularly limited (e.g., each class only has four labels). 2) An extra forward propagation is essential to dynamically regulate $k$ at each iteration, leading to a dramatic drop in the efficiency of model training. 

% \begin{algorithm}[t]
% \caption{FullMatch algorithm.}
% \label{algo fullmatch}
% \begin{algorithmic}[1]
% \STATE \textbf{Input:} $\mathcal{X}=\{(x_m, y_m):m\in(1,...,M)\}$, $ \mathcal{U}=\{\mu_n: n\in(1,...,N)\}$, $\tau$ is the confidence threshold. \{M labeled data and N unlabeled data\};
% \WHILE{not reach the maximum iteration}
% \STATE Generate $P^{(i)}$ and  $Q^{(i)}$\{Predictions of strongly-augmented and weakly-augmented version\};
% \STATE Calculate $k$ using Eq. (\ref{cal k}) \{Generate negative pseudo-labels\};
% \STATE Calculate ANL loss $\mathcal{L}_{anl}$ using Eq. (\ref{cal CNL}) for all unlabeled examples;
% \IF{$max\ (Q^{(i)})\ \geq\ \tau$}
% \STATE Calculate unsupervised loss $\mathcal{L}_{us}$ using Eq. (\ref{cal unsup_loss}) ;
% \STATE Calculate $y_c$ using Eq. (\ref{cal confused labels}) \{ Determine the label of non-target categories\};
% \STATE Calculate EML $\mathcal{L}_{eml}$ using Eq. (\ref{cal em});
% \ELSE
% \STATE $\mathcal{L}_{eml} = 0$, $\mathcal{L}_{us} = 0.$ \{Ignore the examples without pseudo-labels\};
% \ENDIF
% \STATE Calculate supervised loss $\mathcal{L}_{s}$ using Eq. (\ref{cal sup});
% \STATE Update model via $\mathcal{L}_{sum}=\mathcal{L}_{s}+\mathcal{L}_{us}+\mathcal{L}_{anl}+\mathcal{L}_{eml}$;
% \ENDWHILE
% \RETURN{Model parameters}
% \end{algorithmic}
% \end{algorithm}

In this work, we present a scheme to approximately evaluate the top-$k$ performance, referred to as Adaptive Negative Learning (ANL). ANL does not require an additional labeled validation set, nor redundant inference processes. This is inspired by UDA~\cite{ref_uda} that by optimizing the consistency between two augmented versions, the model becomes smoother with respect to changes in the input space and thus, the overall performance can be better. Therefore, our key assumption is that the model performance can be reflected by the consistency of the  predictions with different augmented inputs.  
% That is, for all unlabeled data, we first compute the model's predicted $temp$ labels  given the unlabeled example with a weakly-augmented version,  and then calculate the minimum $k$ depending on the strong-augmented predictions so that accuracy is equal to 1. 
That is, we first compute the $temp$ labels according to the weakly-augmented prediction regardless of whether the max score is larger than the threshold, then we view the $temp$ labels as the ground truth of the strongly-augmented version and calculate the minimum $k$ so as to its top-$k$ accuracy is 100\%. 
This can be formulated as: 
\begin{equation}
k = \mathop{\arg\min_{\theta\in{[2,C]}}} (Acc(P_{t},\hat{Q_{t}}, \theta) = 100\%) \label{cal k}  %p_{m,t}(y|\phi(\mu^{(i)}))
\end{equation}
%and $q_{b}=p_{m}(y|\omega(\mu_{b})$ denotes the prediction distribution with weakly-augmented inputs.
where $\hat{Q_{t}}=\arg\max{(Q,t)}$ is $temp$ labels at step $t$, $P_{t}$ denotes the prediction vector of strongly-augmented and they are calculated across the total batch samples. $Acc$ and $C$ represent the function of calculating top-$k$ accuracy and the number of categories, respectively. Since there are always certain examples without pseudo-labels at each training step (see Fig.~\ref{visualize}\subref{visualize of eml}) and meanwhile we calculate $k$ on $all$ unlabeled data, the over-fitting issue can be alleviated. Finally, we assign negative pseudo-labels to categories ranked after top-$k$ on the prediction distributions of the weakly-augmented version.  As a result, the vector $S^{(i)}$ (Sec. \ref{sec_eml}) can be recalculated as:
\begin{equation}
s_c^{(i)} = \mathbbm{1}[q_c^{(i)}\geq \tau] + \mathbbm{1}[Rank(q_c^{(i)})>k]
\end{equation}
where $Rank$ is a category sorting function based on confidence scores in the descending order. In the early stage of the training process, when the model is fed with different augmented versions on the same samples, the output distribution is significantly different, thus the value of $k$ will be enlarged, i.e., ANL will not afford any negative pseudo-labels when $k=C$. With the optimization of consistency loss (i.e., cross-entropy loss), the model has stronger output invariant to input noise, the value of $k$ will turn smaller, and more negative pseudo-labels will be selected. The adaptive negative learning loss $\mathcal{L}_{anl}$ can be formulated as: 
\begin{equation}
- \frac{1}{B} \sum_{i=1}^{B} \sum_{c=1}^{C} \mathbbm{1}[Rank(q_c^{(i)})>k]\  log(1-p_c^{(i)}) \label{cal CNL}
\end{equation}

% \begin{figure}[t]
% \centering
% \includegraphics[width=1.\linewidth]{image/top-k_acc.pdf}
% \caption{ The top-5 and top-9 accuracy curves of FixMatch during training on CIFAR-10 with 40 label samples. The top-5 prediction is almost correct when iterations are greater than 200k. } \label{topk_acc}
% \end{figure}

Note that the expense of employing ANL is almost free. It does not introduce any extra forward propagation processes for evaluating the performance, nor any new hyper-parameters.  
% Furthermore, unlike UPS~\cite{ref_in_defense} and NS$^3$L~\cite{ref_n3l} that obtain negative pseudo-labels by setting a predefined fixed threshold, our ANL adaptively allocates negative pseudo-labels based on the learned selection strategy (i.e., top-$k$) while maintaining simplicity. 
Unlike UPS~\cite{ref_in_defense} and NS$^3$L~\cite{ref_n3l}, ANL do not rely on the confidence values and allow to allocate negative pseudo-labels to examples  with ambiguous prediction.
For more analysis about ANL (e.g., training with limited labeled examples), please refer to the \textit{Supplementary Material}, Section B.

\begin{table*}[t]
    \renewcommand\arraystretch{1.1}
    \centering
    \setlength{\tabcolsep}{0.65mm}{
    \begin{tabular}{c|ccc|ccc|cc|c}

    \multicolumn{1}{c|}{} & \multicolumn{3}{c|}{\small CIFAR-10} & \multicolumn{3}{c|}{\small CIFAR-100} & \multicolumn{2}{c|}{\small SVHN} & \multicolumn{1}{c}{\small STL-10} \\
    \small Label Amount & \small 40  & \small 250 & \small 4000 & \small 400 & \small 2500 & \small 10000 & \small 40 & \small 1000 & \small 1000 \\
    \Xhline{3\arrayrulewidth}

    \small UDA ~\cite{ref_uda} & \small 89.38\scriptsize$\pm3.75$ & \small 94.84\scriptsize$\pm0.06$ & \small 95.71\scriptsize$\pm0.07$ & \small 53.61\scriptsize$\pm1.59$ & \small 72.27\scriptsize$\pm0.21$ & \small 77.51\scriptsize$\pm0.23$ & \small 94.88\scriptsize$\pm4.27$ & \small \textbf{98.11}\scriptsize$\pm0.01$ & \small 93.36\scriptsize$\pm0.17$ \\
    
    \small RemixMatch~\cite{ref_remixmatch} & \small 90.12\scriptsize$\pm1.03$ & \small 93.7\scriptsize$\pm0.05$ & \small 95.16\scriptsize$\pm0.01$ & \small 57.25\scriptsize$\pm1.05$ & \small 73.97\scriptsize$\pm0.35$ & \small\textbf{79.98}\scriptsize$\pm0.27$ & \small 75.96\scriptsize$\pm9.13$ & \small 94.84\scriptsize$\pm0.31$ & \small 93.26\scriptsize$\pm0.14$ \\
    
    \small Semco$^\ddagger$~\cite{ref_semco}  & \small 92.13\scriptsize$\pm0.22$ & \small 94.88\scriptsize$\pm0.27$ & \small 96.20\scriptsize$\pm0.08$ & \small 55.89\scriptsize$\pm1.18$ & \small 68.07\scriptsize$\pm0.01$ & \small 75.55\scriptsize$\pm0.12$ &-&-&\small 92.51\scriptsize$\pm0.29$ \\
    
    \small Dash~\cite{ref_dash} & \small 86.78\scriptsize$\pm3.75$ & \small 95.44\scriptsize$\pm0.13$ & \small 95.92\scriptsize$\pm0.06$ & \small 55.24\scriptsize$\pm0.96$ & \small 72.82\scriptsize$\pm0.21$ & \small 78.03\scriptsize$\pm0.14$ & \small 96.97\scriptsize$\pm1.59$ & \small 97.97\scriptsize$\pm0.06$ & \small 92.74\scriptsize$\pm0.40$\\
    
    \small UPS~\cite{ref_in_defense} & \small 94.74\scriptsize$\pm0.29$ & \small 94.89\scriptsize$\pm0.08$ & \small 95.75\scriptsize$\pm0.05$ & \small 58.93\scriptsize$\pm1.66$ & \small 72.86\scriptsize$\pm0.24$ & \small 78.03\scriptsize$\pm0.23$ & - & - & \small 93.98\scriptsize$\pm0.28$\\

    \small AlphaMatch$^\ddagger$~\cite{ref_alphamatch} & \small 91.35\scriptsize$\pm3.38$ & \small 95.03\scriptsize$\pm0.29$ & \small - & \small 61.26\scriptsize$\pm3.13^{\dagger}$ & \small \textbf{74.98}\scriptsize$\pm0.27^{\dagger}$ & \small - & \small 97.03\scriptsize$\pm0.26$ & - & \small 90.36\scriptsize$\pm0.75$\\

    \small CoMatch~\cite{ref_comatch} & \small 93.12\scriptsize$\pm0.92$ & \small 95.10\scriptsize$\pm0.35$ & \small 95.94\scriptsize$\pm0.03$ & \small 59.98\scriptsize$\pm1.11$ & \small 72.99\scriptsize$\pm0.21$ & \small 78.17\scriptsize$\pm0.23$ & - & - & \small 91.34\scriptsize$\pm0.41$\\
    
    \small SimMatch$^{\dagger}$$^\ddagger$~\cite{ref_simmatch} & \small 94.40\scriptsize$\pm1.37$ & \small 95.16\scriptsize$\pm0.39$ & \small 96.04\scriptsize$\pm0.01$ & \small 62.19\scriptsize$\pm2.21$ & \small 74.93\scriptsize$\pm0.32$ & \small 79.42\scriptsize$\pm0.11$ & - & - & -\\

    % \small DoubleMatch~\cite{ref_doublematch} & \small 86.41\scriptsize$\pm5.60$ & \small 94.44\scriptsize$\pm0.42$ & \small 95.35\scriptsize$\pm0.17$ & \small 58.17\scriptsize$\pm1.22$ & \small 72.93\scriptsize$\pm0.26$ & \small 78.78\scriptsize$\pm0.17$ & \small 84.63 \scriptsize$\pm11.81$ &  \small 97.9\scriptsize$\pm0.07$ & \small \textbf{95.65}\scriptsize$\pm0.2^{\ddagger}$\\

    \small CR~\cite{ref_cr} & \small 94.31\scriptsize$\pm0.9$ & \small 94.96\scriptsize$\pm0.3$ & \small 95.84\scriptsize$\pm0.13$ & \small 50.77\scriptsize$\pm0.79$ & \small 72.42\scriptsize$\pm0.37$ & \small 78.97\scriptsize$\pm0.23$ & \small 96.33\scriptsize$\pm1.84$ & \small 97.61\scriptsize$\pm0.06$ & \small 93.04\scriptsize$\pm0.42$ \\

    \small NP-Match~\cite{ref_npmatch} & \small 95.09\scriptsize$\pm0.04$ & \small 95.04\scriptsize$\pm0.06$ & \small 95.89\scriptsize$\pm0.02$ & \small 61.08\scriptsize$\pm0.99$ & \small 73.97\scriptsize$\pm0.26$ & \small 78.78\scriptsize$\pm0.13$ & - & - & \small 94.41\scriptsize$\pm0.24$\\

    \hline
    
    \small FixMatch ~\cite{ref_fixmatch} & \small 92.53\scriptsize$\pm0.28$ & \small 95.14\scriptsize$\pm0.05$ & \small 95.79\scriptsize$\pm0.08$ & \small 57.45\scriptsize$\pm1.76$ & \small 71.97\scriptsize$\pm0.16$ & \small 77.8\scriptsize$\pm0.12$ & \small 96.19\scriptsize$\pm1.18$ & \small \textbf{98.04}\scriptsize$\pm0.03$ & \small 93.75\scriptsize$\pm0.33$ \\
    
    \small FullMatch (ours)& \small {\textbf{94.11}}\scriptsize$\pm1.01$ & \small \textbf{95.36}\scriptsize$\pm0.12$ & \small \textbf{96.25}\scriptsize$\pm0.08$ & \small {\textbf{59.42}}\scriptsize$\pm1.40$ & \small {\textbf{73.06}}\scriptsize$\pm0.40$ & \small {\textbf{78.56}}\scriptsize$\pm0.10$ & \small \textbf{97.65}\scriptsize$\pm0.10$ & \small 98.01\scriptsize$\pm0.03$ & \small {\textbf{94.26}}\scriptsize$\pm0.09$\\

    % \small \textbf{$\Delta$} & \small \textcolor{PineGreen}{+1.58} & \small \textcolor{PineGreen}{+0.22} & \small \textcolor{PineGreen}{+0.46} & \small \textcolor{PineGreen}{+1.97} & \small \textcolor{PineGreen}{+1.09} & \small \textcolor{PineGreen}{+0.76} & \small \textcolor{PineGreen}{+1.46} & \small {-0.03} & \small \textcolor{PineGreen}{+0.51}\\
    \hline
    
    \small FlexMatch ~\cite{ref_flexmatch} & \small 95.03\scriptsize$\pm0.06$ & \small 95.02\scriptsize$\pm0.09$ & \small 95.81\scriptsize$\pm0.01$ & \small 60.06\scriptsize$\pm1.62$ & \small 73.51\scriptsize$\pm0.2$ & \small 78.1\scriptsize$\pm0.15$ & \small 96.08\scriptsize$\pm1.24$ & \small 97.37\scriptsize$\pm0.06$ & \small 94.23\scriptsize$\pm0.18$ \\
    
    \small $FullFlex$ (ours) & \small {{\textbf{95.56}}}\scriptsize$\pm0.15$ & \small {{\textbf{95.61}}}\scriptsize$\pm0.04$ & \small \textbf{{96.28}}\scriptsize$\pm0.03$ & \small {{\textbf{62.60}}}\scriptsize$\pm0.64$ & \small{\textbf{74.60}}\scriptsize$\pm0.42$ & \small{\textbf{79.26}}\scriptsize$\pm0.21$ & \small{\textbf{97.48}}\scriptsize$\pm0.04$ & \small {\textbf{97.58}}\scriptsize$\pm0.02$ & \small {{\textbf{94.50}}}\scriptsize$\pm0.12$ \\

    % \small \textbf{$\Delta$} & \small \textcolor{PineGreen}{+0.53} & \small \textcolor{PineGreen}{+0.59} & \small \textcolor{PineGreen}{+0.47} & \small \textcolor{PineGreen}{+2.54} & \small \textcolor{PineGreen}{+1.09} & \small \textcolor{PineGreen}{+1.16} & \small \textcolor{PineGreen}{+1.4} & \small \textcolor{PineGreen}{+0.21} & \small \textcolor{PineGreen}{+0.27}\\
    
    %\bottomrule
    \end{tabular}}
    \caption{\textbf{Top-1 accuracy ($\%$) for CIFAR-10/100, SVHN and STL-10 datasets on 3 different folds.} $FullFlex$ indicates applying our method to FlexMatch. $\dagger$ indicates introducing an additional technique named DA (Distribution Alignment)~\cite{ref_remixmatch}. $\ddagger$ represents the result comes from the original paper.}
    
    \label{table cifar}
\end{table*}

\subsection{FullMatch}

By integrating the Entropy Meaning Loss (EML) and Adaptive Negative Learning (ANL) into FixMatch, we propose an advanced SSL algorithm named FullMatch. Since ANL can allocate negative pseudo-labels for all unlabeled data, it encourages us to take negative pseudo-labels as additional targets into account when calculating $y_{c}^{(i)}$ and $\mathcal{L}_{eml}$. This means the count of non-target class in examples with pseudo-label is $k-1$ instead of $C-1$. 

As shown in Fig.~\ref{FullMatch}, we first calculate $k$ to assign negative pseudo-labels for $all$ unlabeled examples and use Eq.(\ref{cal CNL}) to train the model. This means the network can learn from the low-confidence examples instead of neglecting them directly. Then, we render pseudo-labels based on the prediction of weakly-augmented examples and use cross-entropy as loss function, similar to FixMatch. For examples with pseudo-label, we further view the remaining categories as non-target classes and utilize EML to train the corresponding class outputs in strongly-augmented predictions. Therefore, we can formulate the overall loss in FullMatch as a simple weighted sum of the FixMatch loss (i.e., supervised loss, unsupervised loss), ANL loss and EML: 
\begin{equation}
\mathcal{L}_{sum} = \mathcal{L}_{s} + \mathcal{L}_{us} + \alpha \cdot \mathcal{L}_{anl} + \beta \cdot \mathcal{L}_{eml}
\end{equation}

For simplicity, we set $\alpha$ and $\beta$ to 1. $\mathcal{L}_{s}$ and $\mathcal{L}_{us}$ (Eq. (\ref{cal unsup_loss})) are respectively the supervision loss for labeled samples and the consistency loss for unlabeled samples: 
\begin{equation}
\mathcal{L}_{s} = \frac{1}{B_l} \sum_{i=1}^{B_l} H(y^{(i)}, p_{m}(y|\omega(x^{(i)})) \label{cal sup}
\end{equation}
%section A for the full algorithm of FullMatch and 
where $B_l$ is the batch size of labeled examples. See \textit{Supplementary Material}, Section C.1 for the full algorithm of FullMatch.

\section{Experiments}
We evaluate the efficacy of the proposed FullMatch on several popular SSL datasets: CIFAR-10/100~\cite{ref_cifar}, SVHN~\cite{ref_svhn}, STL-10~\cite{ref_stl10} and ImageNet~\cite{ref_imagenet}, and perform extensive experiments across various amounts of labeled data. In addition, we conduct experiments based on the FlexMatch~\cite{ref_flexmatch} algorithm to further show our method is orthogonal to FlexMatch, which is a strengthened version of FixMatch by introducing $Curriculum$ $Pseudo$ $Labeling$ (CPL) to adjust the threshold dynamically. %and achieve the state-of-the-art performance. 
We also present the ablation study to better understand why our method is effective. 

For fair comparisons, we keep the same hyperparameters as FixMatch and FlexMatch. Specifically, we employ a cosine learning rate decay schedule~\cite{ref_coslr} and standard stochastic gradient descent (SGD) with a momentum of 0.9 as optimizer~\cite{ref_sgd1} across all amounts of labeled examples and datasets, the initial learning rate is 0.03 and the total iteration number is set to $2^{20}$. %These settings follow the original paper. 
We use RandAugment~\cite{ref_randaug} as the strong augmentation in all experiments, and ResNet-50~\cite{ref_resnet} for ImageNet datasets and Wide ResNet~\cite{ref_wrn,ref_wrn2} (e.g., WRN 28-2 and WRN 28-8) for other benchmarks. Our framework is implemented on TorchSSL~\cite{ref_flexmatch}. See \textit{Supplementary Material}, Section C.2 for details.

\subsection{Main Results}

We report the performance of FullMatch on the four popular SSL benchmarks: CIFAR-10/100, SVHN and STL-10, as shown in Table~\ref{table cifar}. We calculate the mean and variance of top-1 accuracy when training on 3 different ``folds'' of labeled data. The results of other algorithms are mainly from TorchSSL and NP-Match~\cite{ref_npmatch}. We use few new results for a fair comparison (e.g., the performance of FlexMatch on SVHN), which are better than their published version. The experiments show that FullMatch outperforms FixMatch under all amounts of labeled data and benchmarks, except the SVHN dataset with 1000 labels. Additionally, since our method can be integrated with any variants based on FixMatch, we also use the state-of-the-art method FlexMatch~\cite{ref_flexmatch} as the baseline and called the integrated method $FullFlex$. The results demonstrate that our method can significantly boost the baseline models, including FixMatch and FlexMatch for almost all datasets, and meanwhile surpasses the latest methods, e.g., NP-Match~\cite{ref_npmatch}, SimMatch~\cite{ref_simmatch} and DoubleMatch~\cite{ref_doublematch}. Our method has the following advantages: 

1) FullMatch brings considerable improvements compared with FixMatch, especially when the amount of labeled data is extremely limited. We report the performance of different algorithms with only 4 labels per class, corresponding to the 40 labeled data of CIFAR-10, SVHN and 400 labeled data of CIFAR-100. The results indicate the average accuracy of FullMatch exceeds FixMatch by more than 1\%, especially on CIFAR-100 (an increase of 2\%).

2) FullMatch is efficient. Our method does not introduce any extra hyperparameters or extra forward propagation process. Fig.~\ref{inference_time} compares the average training time cost of a single iteration with and without using our method on Geforce RTX 2080 Ti GPU. It is obvious that while enhancing the performance of existing algorithms, our method introduces negligible computational overhead. 

3) The proposed FullMatch is orthogonal to existing popular methods. Namely, our method can further improve the performance of other  FixMatch-based methods. For instance, we introduce CPL into FullMatch, named $FullFlex$. The extensive experiments show that FullFlex achieves state-of-the-art performance under almost all benchmarks. 
%except for SVHN. We emphasize the fact that FlexMatch itself is slightly weaker than FixMatch on SVHN. 

\subsection{Results on ImageNet}

We also evaluate our method on ImageNet to confirm the effectiveness on a more realistic and larger dataset. Following TorchSSL settings, we select samples with 100 labels for each class, which is less than 8\% of the total train set. 
% Note we use the same seed and rules to ensure the labeled data are the same. 
Table~\ref{tab imagenet} shows the performance of different algorithms after $2^{20}$ iterations, where the results of different methods come from TorchSSL and NP-Match~\cite{ref_npmatch} (it conducts all experiments based on TorchSSL settings). When all hyper-parameters are kept consistently with TorchSSL, the top-1 accuracy of FullMatch and FullFlex are both improved by more than 1\%, which further confirms the effectiveness of our method on this complicated dataset. Furthermore, compared with the latest methods, our method still achieve a state-of-the-art performance.

% Due to the computational resource limitation, many SSL methods have not been evaluated on ImageNet (e.g., Dash). Note that the reported results may be not the best performance of each method, and owing to the computational resource burden, we did not intentionally tune any hyperparameters. But it still indicates the effectiveness of our proposed method.

\begin{table}[!htbp]
    \centering
    \renewcommand\arraystretch{1.03}
    \centering
    \setlength\tabcolsep{7pt}{
    \begin{tabular}{c|cc}
    %\toprule
    \small  & \small Top-1  & \small Top-5 \\
    \Xhline{3\arrayrulewidth}
    % \footnotesize CoMatch & \footnotesize 57.83  & \footnotesize 80.36 \\
    \small UPS~\cite{ref_in_defense} & \small 57.31  & \small 79.77 \\
    \small NP-Match~\cite{ref_npmatch} & \small 58.22  & \small 80.67 \\
    \hline
    \small FixMatch~\cite{ref_fixmatch} & \small 56.34  & \small 78.20 \\
    \small FullMatch (ours) & \small \textbf{57.44} (\small \textcolor{PineGreen}{+1.1}) & \small \textbf{79.26} (\textcolor{PineGreen}{+1.06}) \\
    %\small \textbf{$\Delta$ }&  & \small  \\
    \hline
    \small FlexMatch~\cite{ref_flexmatch}& \small 58.15  & \small 80.52 \\
    \small $FullFlex$ (ours) & \small \textbf{59.58} (\textcolor{PineGreen}{+1.43}) & \small \textbf{81.38} (\textcolor{PineGreen}{+0.86}) \\
    % \small \textbf{$\Delta$ }& \small \textcolor{PineGreen}{+1.43} & \small \textcolor{PineGreen}{+0.86} \\
    \end{tabular}}
    \caption{\textbf{Top-1 and Top-5 accuracy (\%) on ImageNet.}  In green are the values of performance improvement over the baselines.}
    \label{tab imagenet}
    \vspace{-3pt}
\end{table}

\subsection{Ablation Study}

Since FullMatch is essentially a combination of two novel techniques and FixMatch, we present an ablation study to verify the effectiveness of different components. 

~\textbf{Entropy Meaning Loss (EML)}. We conduct an ablation study on EML, as shown in Table~\ref{ablation study}.  It can be seen that FixMatch can obtain obvious improvement when combining EML. We also conduct experiments with different implementations of EML, seeing row.3 vs row.4, we can find that the proposed EML is very effective, regardless of the form of the loss function.

~\textbf{Adaptive Negative Learning (ANL)}. We study the role of ANL for the model. If we only afford negative pseudo-labels to examples with pseudo-labels, there will only be a slight improvement (57.83 vs 57.68), see Table~\ref{ablation study}. It is believed that the slight gain mainly comes from alleviating the negative effect of incorrect pseudo-labels because negative pseudo-labels are always less noisy. Moreover, if we assign negative pseudo-labels to the samples that are without positive pseudo-labels, it can bring a considerable performance gain (0.91\%). When we assign all data the negative pseudo-labels, the performance is further improved (1\%), indicating that our model can extract discriminative knowledge from all unlabeled data. 
Finally, if we apply EML and ANL to FixMatch, the FullMatch exceeds the baseline with a large margin (1.64\%), which shows that the proposed two components are useful and complementary. % and can learn from each other for the best performance. 

\begin{table}[t]
    \renewcommand\arraystretch{1.1}
    \centering
    \setlength{\tabcolsep}{1.8mm}{
    \begin{tabular}{c|cccc|cc}
     \small  & \small CE & \small BCE & \small w PL & \small w/o PL & \small Accuracy & \small \textbf{$\Delta$}\\
        \Xhline{3\arrayrulewidth}
     \small FixMatch&  &  &  &  & \small 57.68 & -\\
    \hline
    \multirow{2}{*}{\small EML}
    & \checkmark &  &  &  & \small 58.35 & \textcolor{PineGreen}{+0.67}\\
    &  & \checkmark &  &  & \small \textbf{58.47} & \textcolor{PineGreen}{+0.79}\\
    
    \hline
    \multirow{3}{*}{\small ANL}
    & &  & \checkmark &  & \small 57.83 & \textcolor{PineGreen}{+0.15}\\
    &  &  &  & \checkmark & \small 58.59 & \textcolor{PineGreen}{+0.91}\\
    &  &  & \checkmark & \checkmark & \small \textbf{58.67} & \textcolor{PineGreen}{+0.99}\\
    \hline
    \small FullMatch &  &\checkmark & \checkmark &\checkmark & \small \textbf{59.32} & \textcolor{PineGreen}{+1.64} \\
    %\bottomrule
    \end{tabular}}
    \caption{\textbf{Ablation study of FullMatch on 400-label split from CIFAR-100.} CE and BCE represent the loss implementation of EML. ``w PL" and ``w/o PL" means applying ANL on examples with/without pseudo-label, respectively. \textbf{$\Delta$} represents the performance improvement over the baseline.}
    \vspace{-3pt}
    \label{ablation study}

\end{table}

\textbf{Different $\alpha$ and $\beta$}. Table~\ref{tab alpha_beta} reports the accuracy (\%) of different $\alpha$ and $\beta$. It reveals FullMatch is not sensitive to $\alpha$ and $\beta$. We take untuned weights (i.e. $\alpha, \beta=1$ ) for all experiments to show the gains come entirely from our method.

\begin{table}[!htbp]
    \centering
    \renewcommand\arraystretch{1.}
    \centering
    \setlength\tabcolsep{2pt}{
    \begin{tabular}{c|ccc|ccc|ccc}
    
    \multicolumn{1}{c|}{\small $\alpha$} & \multicolumn{3}{c|}{\small 0.5} & \multicolumn{3}{c|}{\small 1.0} & \multicolumn{3}{c}{\small 2.0}\\

    \small $\beta$ & \small 0.5 & \small 1.0 & \small 2.0 & 
    \small 0.5 & \small 1.0 & \small 2.0 & \small 0.5 & \small 1.0 & \small 2.0 \\
    
    \Xhline{3\arrayrulewidth}
    
    \small $acc$ & \small 78.43 & \small 78.36 & \small 78.50 & 
    \small 78.38 & \small 78.46 & \small 78.48 & \small 78.49 & \small 78.31 & \small 78.47 \\

    \end{tabular}}
    \caption{\textbf{Ablation study on $\alpha$ and $\beta$.} All experiments are conducted on CIFAR-100 with 10000-label.}
    \vspace{-3pt}
    \label{tab alpha_beta}
    
\end{table}

\section{Conclusion}

In this paper, we first analyze the unlabeled data wasting in the FixMatch-based methods, and then we are motivated to propose two novel techniques: Entropy Meaning Loss (EML) and Adaptive Negative Learning (ANL). The EML explicitly constrains the output distribution of non-target classes to produce more high-confidence predictions, thus selecting more examples with pseudo-label under the same threshold. The ANL introduces additional negative pseudo-labels to learn knowledge from low-confidence examples without pseudo-label. Note that ANL assesses the top-$k$ performance dynamically to allocate negative pseudo-labels and does not introduce any extra hyper-parameters. FullMatch, the proposed method based upon FixMatch, achieves significant improvement in most scenarios while being extremely concise and efficient. In addition, we also integrate our method with FlexMatch, which achieves state-of-the-art performance on a variety of SSL benchmarks. Experimental results strongly confirm the effectiveness of our method. We believe this work will provide new insights to explore the low-confidence unlabeled data in SSL. 

%%%%%%%%% REFERENCES
{\small
\bibliographystyle{ieee_fullname}
\bibliography{egbib}
}

\end{document}